%% file: main.tex
\documentclass[10pt,twocolumn,letterpaper]{article}

\usepackage{icb}
\usepackage{times}
\usepackage{epsfig}
\usepackage{graphicx}
\usepackage{amsmath}
\usepackage{amssymb}
\usepackage{color}
\usepackage{booktabs}
\usepackage{caption}
\usepackage{subcaption}
\usepackage{multirow}
\usepackage{lipsum}

\makeatletter
\def\blfootnote{\gdef\@thefnmark{}\@footnotetext}
\makeatother

\usepackage[pagebackref=true,breaklinks=true,colorlinks,bookmarks=false]{hyperref}

\icbfinalcopy 

\ificbfinal\fi
\begin{document}

\title{RoPAD: Robust Presentation Attack Detection through \\ Unsupervised Adversarial Invariance}

\author{Ayush Jaiswal\textsuperscript{*}, Shuai Xia\textsuperscript{*}, Iacopo Masi, Wael AbdAlmageed\\
USC Information Sciences Institute, Marina del Rey, CA, USA\\
{\tt\small \{ajaiswal,sxia,iacopo,wamageed\}@isi.edu}
}

\maketitle
\thispagestyle{empty}

\begin{abstract}
For enterprise, personal and societal applications, there is now an increasing demand for automated authentication of identity from images using computer vision. However, current authentication technologies are still vulnerable to presentation attacks. We present RoPAD, an end-to-end deep learning model for presentation attack detection that employs unsupervised adversarial invariance to ignore visual distractors in images for increased robustness and reduced overfitting. Experiments show that the proposed framework exhibits state-of-the-art performance on presentation attack detection on several benchmark datasets.\blfootnote{\textsuperscript{*}Ayush Jaiswal and Shuai Xia contributed equally.}
\end{abstract}


\input{./sections/01_introduction}

\input{./sections/02_related_work} 

\input{./sections/03_ropad}

\input{./sections/04_eval}

\input{./sections/05_conclusions}

\section*{Acknowledgements}

This research is based upon work supported by the Office of the Director of National Intelligence (ODNI), Intelligence Advanced Research Projects Activity (IARPA), via IARPA R\&D Contract No. 2017-17020200005. The views and conclusions contained herein are those of the authors and should not be interpreted as necessarily representing the official policies or endorsements, either expressed or implied, of the ODNI, IARPA, or the U.S. Government. The U.S. Government is authorized to reproduce and distribute reprints for Governmental purposes notwithstanding any copyright annotation thereon.

{\small
\bibliographystyle{ieee}
\bibliography{main}
}
\end{document}

%% file: sections/01_introduction.tex
\section{Introduction}
\label{sec:introduction}
\begin{figure}[t]
    \centering
    \begin{subfigure}{0.48\linewidth}
    \captionsetup{belowskip=2pt}
    \centering
    \includegraphics[width=\linewidth]{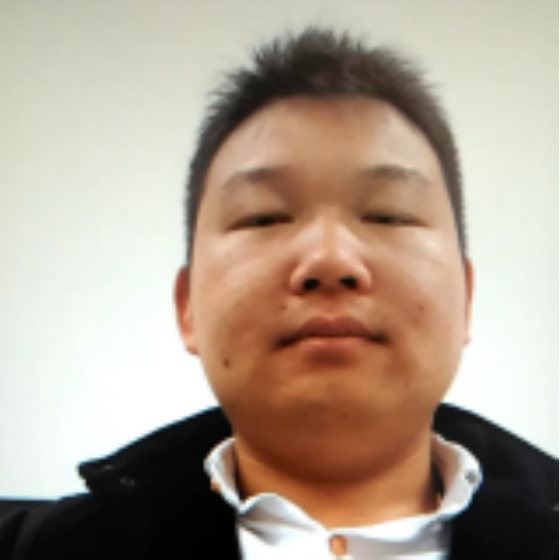}
    \caption{}\label{teaser_a}
    \end{subfigure}
    \begin{subfigure}{0.48\linewidth}
    \captionsetup{belowskip=2pt}
    \centering
    \includegraphics[width=\linewidth]{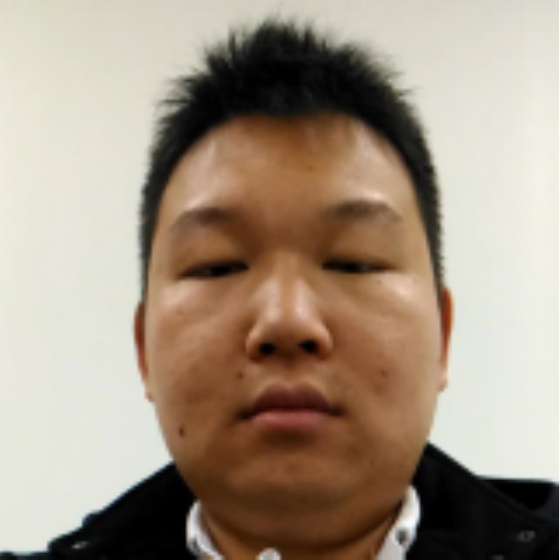}
    \caption{}\label{teaser_b}
    \end{subfigure}
    \begin{subfigure}{0.48\linewidth}
    \centering
    \includegraphics[width=\linewidth]{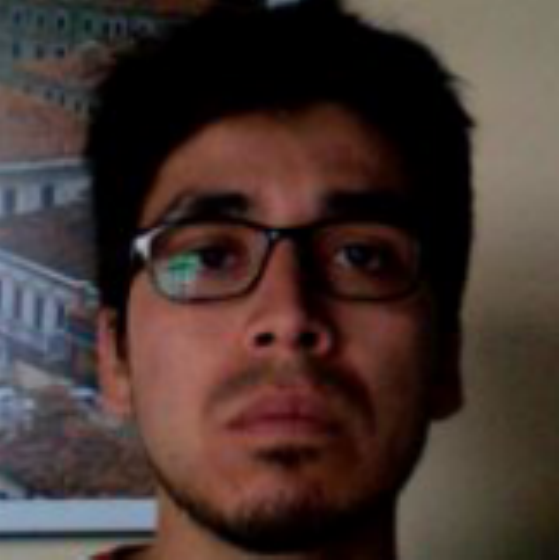}
    \caption{}\label{teaser_c}
    \end{subfigure}
    \begin{subfigure}{0.48\linewidth}
    \centering
    \includegraphics[width=\linewidth]{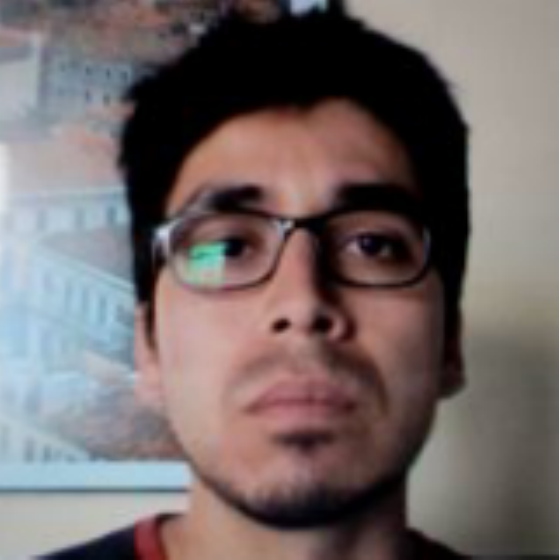}
    \caption{}\label{teaser_d}
    \end{subfigure}
    \caption{Among all the images, which are genuine images and which are presentation attacks?\protect\footnotemark}
    \label{teaser}
\end{figure}

Biometric identity authentication technologies based on computer vision, such as face and iris recognition, have become ubiquitous in recent times. However, biometric authentication methods are still prone to presentation attacks, where spoof samples (e.g. printed pictures or videos of a person) are presented to the biometric sensor, attempting to gain unauthorized access. Furthermore, other factors, such as the rising ease of 3D printing technology and capturing very realistic high-resolution images and videos of people's faces due to advancements in camera technologies as well as generative adversarial networks make creating these presentation attacks much easier. As illustrated in Figure~\ref{teaser}, which shows samples of genuine faces and presentation attacks, learning subtle feature to differentiate the two is very challenging even for humans. Therefore, it is crucial to augment face recognition systems with presentation attack detection (PAD) methods in order to improve the security of face authentication systems.

Presentation attack detection methods can be broadly categorized into two classes. The first class of methods depends on augmenting the biometric authentication hardware with an additional sensor that provides auxiliary data that can be used (with or without the original biometric data) by a presentation attack detection algorithm. 
For example, light field cameras (LFCs) have been used to capture multiple depth images of faces, which are then analyzed through a rule-based scheme for PAD~\cite{raghavendra2015presentation}. This class of methods is limited by large cost and legacy system compatibility constraints. The second class of methods  directly uses regular data captured by the authentication system for presentation attack detection using, for example,  signal processing and/or machine learning algorithms. These  methods\footnotetext{\rotatebox{180}{(b) and (c) are genuine faces for authentication.}} extract features, such as Local Binary Patterns (LBP)~\cite{ramachandra2017presentation}, and classify them as bona fide or attack using a downstream classifier, such as support vector machine, or use a convolutional neural network (CNN) for both learned representation extraction and classification~\cite{jourabloo2018face}. The second class of approaches for PAD are, however, inherently challenging and have garnered tremendous research interest in recent times. While the use of deep neural networks (DNNs) in PAD has led to major boosts in performance~\cite{jourabloo2018face}, their inherent limitations, such as vulnerability to overfitting, need for vast amounts of training data, etc. prevent DNN-based systems from reaching their full potential.

One such limitation of DNNs, like most machine learning models, is that they could learn incorrect associations between nuisance factors in the raw data and the final prediction target (e.g. pose, gender or skin tone nuisance factors in face recognition), leading to poor generalization. Existing DNN-based PAD methods do not address this inherent problem and can, hence, be made more robust by incorporating learning techniques that induce robustness through invariance to nuisance factors.

In this paper, we propose RoPAD, a novel end-to-end deep neural network model for presentation attack detection that robustly classifies face images as ``live'' (i.e. real) or ``spoof'' (i.e. fake)  by being invariant to visual distractors inherent in images. The invariance is achieved by adopting the unsupervised adversarial invariance (UAI) framework~\cite{jaiswal2018unsupervised}, which induces implicit feature selection and invariance to nuisance factors within neural networks without requiring nuisance annotations. Most of the visual content in face images is not relevant for PAD. For example, given a face image, the identity of the person, variations in the pose of the face, fine-grained facial attributes, and elements of the background of the image are irrelevant to PAD. Therefore, employing UAI as a core component of the proposed model makes it largely invariant to all such distractors in an inexpensive yet effective way.

The proposed RoPAD model exhibits state-of-the-art performance on 3DMAD~\cite{Erdogmus2014Spoofing}, Idiap Replay-Mobile~\cite{costa2016replay}, Idiap Replay-Attack~\cite{chingovska2012effectiveness}, MSU-MFSD~\cite{wen2015face} and GCT1, a new self-collected dataset (described in Section~\ref{GCT1_desc}), which includes common forms of presentation attacks studied in recent literature, viz., printed faces-images, video replays and 3D masks~\cite{Erdogmus2014Spoofing,ramachandra2017presentation}. Ablation study of a base model (BM), which does not include UAI, shows that UAI provides a significant boost in performance. This essentially proves that invariance to visual distractors makes PAD significantly more robust and effective.

The rest of the paper is organized as follows. Section~\ref{sec:related} provides a brief review of existing face PAD methods. In Section~\ref{sec:method} we discuss a data factorization rationale of PAD and describe the proposed RoPAD model. Section~\ref{sec:eval} summarizes the results of our experimental evaluation. Finally, Section~\ref{sec:conclusion} concludes the paper and discusses directions for future work.

%% file: sections/02_related_work.tex
\section{Related Work}
\label{sec:related}

PAD methods for image-based biometric authentication have traditionally involved extraction of discriminative features, such as specular reflection, blurriness,
chromatic moment, color diversity, etc. and their analysis to distinguish live (genuine) images from spoof (fake) ones~\cite{ramachandra2017presentation,wen2015face}. Previous works have also incorporated deep learning based latent features computed offline in conjunction with linear classifiers for PAD as well as  learned representations within a neural network trained end-to-end for the PAD task~\cite{peng2017face}.

Hand-crafted features and statistical machine learning algorithms have been extensively used in the past for the detection of print and replay kind of attacks. For example, texture analysis through extraction of low-level texture features has been widely utilized for spoofing detection~\cite{maatta2011face}. Feature descriptors such as Local Binary Patterns (LBP)~\cite{Gragnaniello2015AnIO}, Scale-Invariant Feature Transform (SIFT)~\cite{Gragnaniello2015AnIO} and Speeded Up Robust Features (SURF)~\cite{boulkenafet2017face} have been popularly employed to embed faces into low dimension encodings in prior works~\cite{liu2018learning}. In order to make such feature descriptors more discriminative for PAD, researchers have utilized different color-spaces such as RGB, HSV and YCbCr~\cite{boulkenafet2015face}. Hand-crafted feature based methods play an important role in the detection of spoofing given their simplicity and effectiveness in PAD for the domains for which they are specifically designed. For instance, several texture analysis methods~\cite{peng2018ccolbp} achieved good performance on the MSU MFSD dataset.

Deep learning has provided powerful approaches for the development of effective data-driven models for a plethora of computer vision tasks. Convolutional Neural Networks (CNNs) have been employed successfully for PAD recently~\cite{nogueira2016fingerprint}. Yang \textit{et al.}~\cite{yang2014learn} were an early adopter of DNNs for PAD, who achieved significant improvements in detection performance with simple CNN architectures. More recently, architectures such as 3D-CNN~\cite{gan20173d}, patch-based and depth-based architectures have been used for PAD~\cite{atoum2017face}. Furthermore, instead of binary supervision, some works utilize spatial and temporal auxiliary information to guide the training of PAD models~\cite{liu2018learning}. 

The proposed RoPAD is a DNN-based model for PAD from raw RGB images that uses a simple CNN architecture coupled with effective unsupervised invariance induction through UAI for robust PAD \emph{without incorporating any of the aforementioned specialized architecture designs or training regimens}.

\begin{figure*}
    \centering
    \begin{subfigure}[t]{\linewidth}
    \centering
    \includegraphics[width=\linewidth]{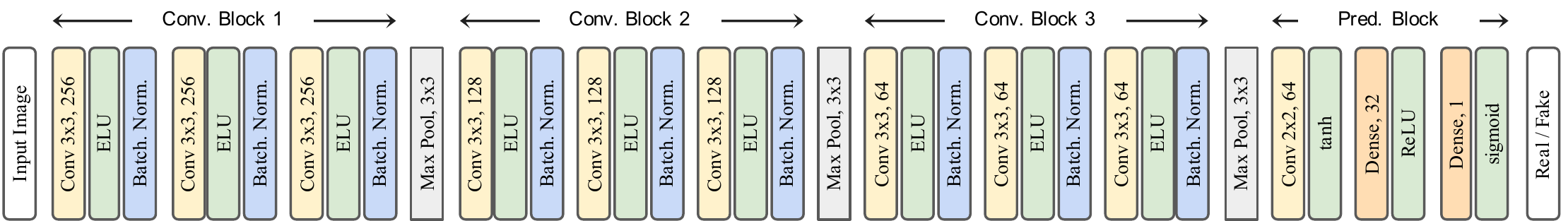}
    \caption{Base CNN Model of RoPAD}\label{base_model}
    \end{subfigure}
    \begin{subfigure}[t]{0.8\linewidth}
    \centering
    \includegraphics[width=\linewidth]{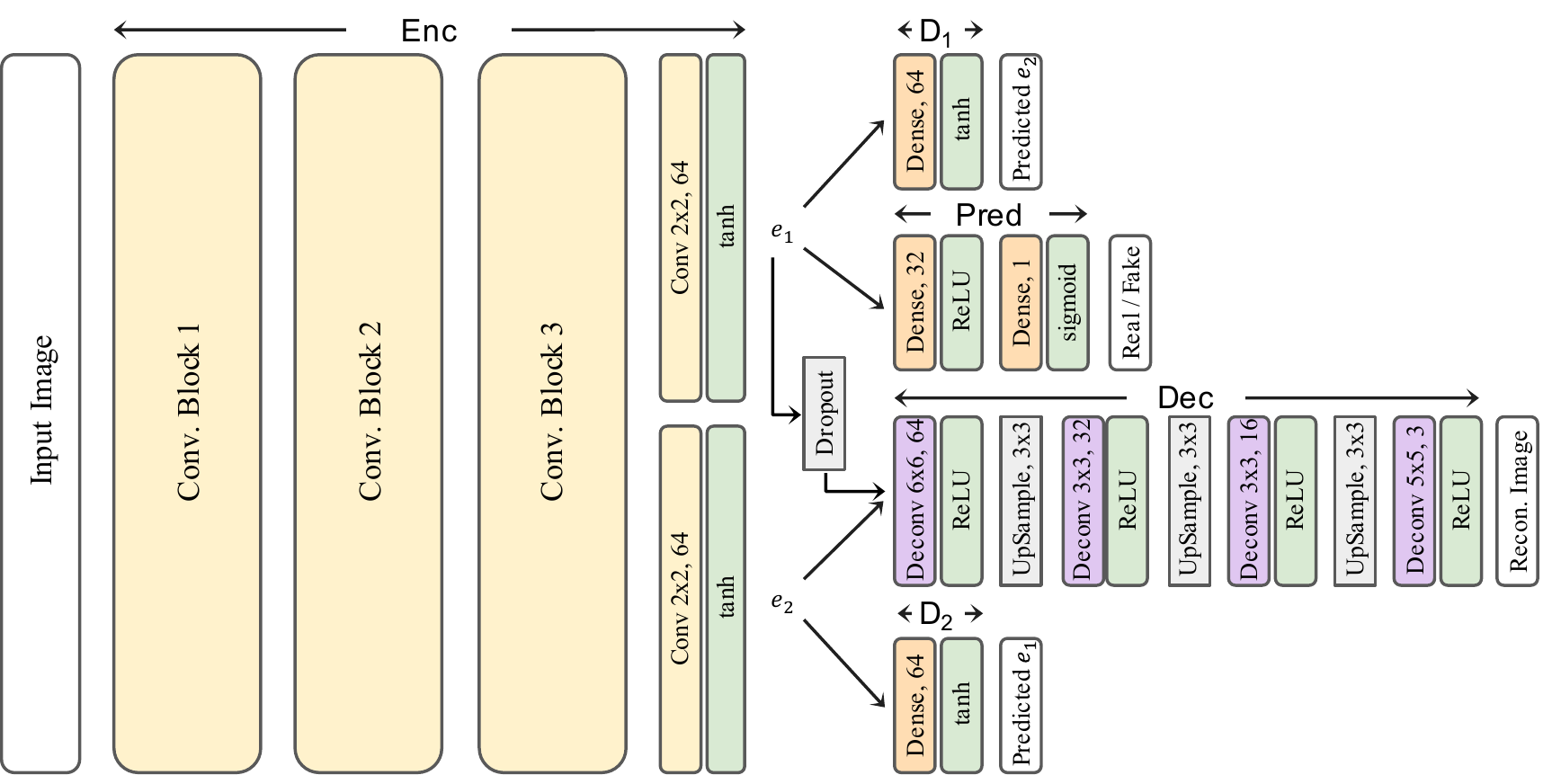}
    \caption{RoPAD training architecture}\label{UAI_figure}
    \end{subfigure}
    \caption{ (a) Base CNN Model of RoPAD: the base model (BM) is inspired by VGG16~\cite{vgg16}, with channel-sizes, activation shapes, activation functions, and batch-normalization being notable modifications. The model comprises three convolutional blocks and a prediction block. (b) RoPAD training architecture: the base model is split into encoder (Enc) and predictor (Pred), and a decoder (Dec) and two disentanglers ($\text{D}_{\text{1}}$ and $\text{D}_{\text{2}}$) are attached for unsupervised adversarial invariance (UAI).}
    \label{base_model_all}
\end{figure*}

%% file: sections/03_ropad.tex
\section{Robust Presentation Attack Detection}
\label{sec:method}

The proposed RoPAD is a DNN model for robust presentation attack detection, which learns to distinguish real face images from fake ones directly from RGB images in an end-to-end framework. Robust PAD is achieved by combining invariant-representation induction and the DNN's ability to learn highly discriminative representations.

\subsection{A Data Factorization View of PAD}\label{sec:problem}
The face image formation process ($\mathcal{S}$) is a complex interaction of multiple entangled signals. For presentation attack detection, we are ultimately interested in the genuineness of a presented sample, and therefore the entangled signals can be split into two main categories --- (1) signals useful for solving the anti-spoofing problem and (2) nuisances for the PAD task. Thus, a face image could be expressed as the result of different factors interacting together as $\mathbf{I} \doteq \mathcal{S}(\boldsymbol{\rho},\boldsymbol{\theta})$, where $\boldsymbol{\rho}$ represents the nuisances defined as all the signals presented in the input media that should not be contributing to the assessment of the genuineness detection, whereas $\boldsymbol{\theta}$ indicates all the signals that are helpful to solving the PAD task. Most common nuisances for PAD tasks can be the subjects's identity, facial attributes, and elements of the background. Contrastively, signals useful for PAD include subtle differences of specific patterns, and characteristic noise affecting non bona fide images. Given all aforementioned variables, presentation attack detection can be improved by reverse-engineering the image formation process $\boldsymbol{\theta}^{\star} = \mathcal{r}\big(\mathcal{S}(\boldsymbol{\theta},\boldsymbol{\rho}) \big)$ to disentangle the important information from the irrelevant ones.

Specifically, given an image $\mathbf{I}$, a PAD system needs to analyze $\boldsymbol{\theta}$, without being distracted by other confounding factors $(\boldsymbol{\rho})$ present implicitly in the media (e.g. identity, pose, background, etc.). Note that, at test time, a PAD system has only access to $\mathbf{I}$ and no access to the individual variables contributing to $\mathcal{S}$ whatsoever.

\subsection{Base CNN Model of RoPAD}\label{Baseline}
The core neural network that is responsible for learning to distinguish between real and fake images is a deep CNN composed of three convolutional blocks and a final prediction block interspersed with max-pooling operations. Each convolutional block contains three convolutional layers alternating with batch-normalization. The kernel shape of each of these convolutional layers is $(3 \times 3)$, and the max-pooling is performed over windows of $(3 \times 3)$. The channel size in each block is kept fixed at $256$, $128$, and $64$ in the first, second, and third blocks, respectively. Exponential Linear Unit (ELU)~\cite{elu} is used as an activation function in the convolutional layers of these blocks. The prediction block contains one convolutional layer with a kernel-shape of $(2 \times 2)$ with the hyperbolic tangent activation, followed by a reshape operation to produce a $64$-dimensional embedding, which is followed by two fully connected layers of output-sizes $32$ and $1$ to perform the final binary classification task (bona fide versus spoof). Figure~\ref{base_model} illustrates the complete architecture. The model design is based on VGG16~\cite{vgg16}, with channel-sizes, activation shapes, activation functions, and batch-normalization being notable modifications, among others. We empirically found the proposed architecture to perform significantly better than the standard VGG16.

\subsection{Unsupervised Adversarial Invariance}
\label{UAI}

Deep neural networks, like machine learning models in general, often learn incorrect associations between the prediction target and nuisance factors of data, leading to poor generalization~\cite{jaiswal2018unsupervised}. Popular approaches to solve this problem include data augmentation controlled for certain nuisance factors or supervised invariance induction to eliminate those factors from the latent representation learned by those models. In both of those approaches, prior knowledge of nuisance factors is necessary. Additionally, the latter approach of invariance induction, which performs significantly better than the former~\cite{jaiswal2018unsupervised} requires annotations of nuisance factors to help guide their elimination from the latent space. \emph{This is especially problematic for PAD because it is a relatively new area of research and suffers from both the lack of expert knowledge about nuisance factors as well as nuisance annotations.} For example, in the case of face images for PAD, the identity of the person, their facial attributes, elements of the background of the image, etc., can be considered nuisance factors for the PAD task. While some of these factors can be annotated with large investments of time and money, others like ``elements of the background'' are difficult to quantify concretely.

Jaiswal \textit{et al.}~\cite{jaiswal2018unsupervised} introduced an unsupervised approach for learning invariance to all, \emph{including potentially unknown}, nuisance factors with respect to a given supervised task. Their unsupervised adversarial invariance (UAI) framework learns a split representation of data into relevant and nuisance factors with respect to the prediction task without needing annotations for the nuisance factors. The underlying mechanism of UAI is formulated as a competition between the prediction and a reconstruction objective coupled with disentanglement between the two representations. This forces the prediction model to utilize only those factors of data that are truly essential for the supervised task at hand (here classificaiton of genuine/fake samples), disregarding everything else.

The UAI framework splits a feedforward neural network into an encoder and a predictor. The encoder is modified such that it produces two representations (i.e. embedding vectors) instead of one -- $e_1$ and $e_2$, where only $e_1$ is used for the prediction task. Besides the encoder and the predictor, the UAI framework consists of a decoder that reconstructs data from a noisy version of $e_1$ concatenated with $e_2$, and a pair of disentanglers that aim to predict one embedding from the other. The disentanglers are trained adversarially against the rest of the model, leading to disentanglement between the two embeddings. The aforementioned competition between the prediction and reconstruction tasks is induced by the noisy channel that connects $e_1$ to the decoder and the enforced disentanglement between $e_1$ and $e_2$, which leads to information separation such that factors of data truly relevant for the prediction task are encoded in $e_1$ and all other factors of data (nuisance) migrate to $e_2$. UAI has been shown~\cite{jaiswal2018unsupervised} to work effectively across a diverse collection of datasets and nuisance factors. Hence, we employ UAI as an integral component in the proposed model.

\begin{figure*}
\begin{center}
 \includegraphics[width=\linewidth]{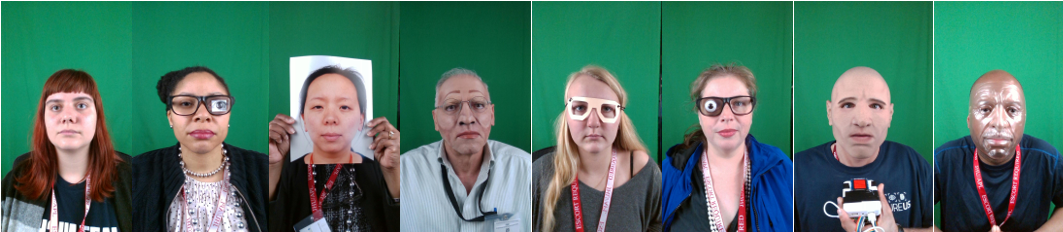}
\end{center}
\caption{\label{GCT1_display}From left to right, corresponding examples of image-types -- (1) genuine, (2) glasses with doll eye, (3) analog photo, (4) makeup, (6) paper glasses, (7) glasses with Van Dyke eye, (8) silicone mask, and (9) transparent mask.}
\end{figure*}

\subsection{RoPAD using UAI}
As mentioned in Section~\ref{UAI}, UAI splits the base feedforward (illustrated in Figure~\ref{base_model}) model into an encoder and a predictor. We split the base CNN model of RoPAD at the prediction block, such that all convolutional blocks as well as the convolutional layer of the prediction block are collectively treated as an encoder, while the two fully connected layers are treated as a predictor. In order to produce two embedding vectors from the encoder instead of one, as required by UAI, we duplicate the final convolutional layer of the encoder as a parallel branch emerging from the final convolutional block. The decoder is designed as a deconvolutional network with four deconvolutional layers interspersed with upsampling layers. The disentanglers are designed as single fully-connected layers, and the noisy transformer is implemented as multiplicative Bernoulli noise, following the approach of~\cite{jaiswal2018unsupervised}. The complete RoPAD architecture is shown in Figure~\ref{UAI_figure}. RoPAD is implemented in Keras\footnote{https://keras.io/} and trained with TensorFlow\footnote{https://www.tensorflow.org/} backend. We follow the same adversarial training strategy as prescribed in~\cite{jaiswal2018unsupervised} of alternating between training the disentanglers versus the rest of the model with $5:1$ frequency.

Performing presentation attack detection with RoPAD at test time is as efficient as with the base model. Although Figure~\ref{base_model_all} shows a complex architecture used for training, RoPAD is actually very light at prediction time. 
At test time, RoPAD is reshaped so that the decoder and the disentangler components of UAI are discarded. Thus, in terms of efficiency and model complexity, RoPAD testing model has the same structure as the base model, yet provides significantly more effective and robust predictions. In light of this, prediction with RoPAD remains as easy as a single forward pass without any additional computational cost.

%% file: sections/04_eval.tex
\section{Experimental Evaluation}
\label{sec:eval}

\subsection{Datasets And Metrics}
\label{GCT1_desc}

\begin{table}
\centering
\setlength{\tabcolsep}{0.375em} 
\begin{tabular}{l ccc}
\toprule
\textbf{Database} & \textbf{Institute} & \textbf{Real/Fake} & \textbf{Attack Types}\\
\cmidrule(r){1-1}  \cmidrule(l){2-4}
3DMAD          & Idiap  &170 / 85  & 3D masks\\
Replay-Attack  & Idiap &  200 / 1000 & printed \& replay\\
Replay-Mobile  & Idiap &  390 / 640  & printed \& replay\\
MSU MFSD       & MSU   & 110 / 330  & printed \& replay\\
\bottomrule
\end{tabular}
\caption{Summary of benchmark datasets for PAD}\label{dataset_desc}
\end{table}

The proposed RoPAD model is evaluated on the following publicly available benchmark datasets for presentation attack detection -- 3DMAD~\cite{Erdogmus2014Spoofing}, Idiap Replay-Mobile~\cite{costa2016replay}, Idiap Replay-Attack~\cite{chingovska2012effectiveness}, and MSU MFSD~\cite{wen2015face}. Details of these datasets are summarized in Table~\ref{dataset_desc}. While 3DMAD exclusively contains presentation attacks involving 3D masks, the other datasets contain attacks through printed faces and video-replays.

RoPAD is also evaluated on the Government Controlled Testing-1 (GCT1) dataset. GCT1 is a dataset collected by Johns Hopkins Applied Physics Laboratory during Government testing of the IARPA Odin project\footnote{Public release of GCT1 is planned by the National Institute of Standards and Technology (NIST)}. GCT1 contains images of about $400$ subjects, including various forms of presentation attacks. The subjects were split into training, validation and testing sets, such that all images of a given subject belonged to only one of the three sets. Table~\ref{GCT1_description} summarizes the attack types and distribution of the attacks in the training, validation and testing sets, which contain $215$, $57$ and $137$ subjects, respectively. We will make the splits publicly available as soon as GCT1 is released by NIST.

\begin{table}
\setlength{\tabcolsep}{0.47em} 
\centering
\begin{tabular}{lccc}
\toprule
\textbf{Type} &\textbf{Train} &\textbf{Validation} &\textbf{Test} \\
\cmidrule(r){1-1}  \cmidrule(l){2-4}
Genuine       &266 &70 &167 \\
Glasses with Doll Eye        &26 &6 &16 \\
Analog Photo       &30  &7  &18 \\
Makeup       &14 &4  &9 \\
Paper Glasses       &27  &7  &17 \\
Glasses with Van Dyke Eye       &26  &7  &17 \\
Silicone Mask       &3  &3  &4 \\
Transparent Mask       &20  &5  &13 \\
\bottomrule
\end{tabular}
\caption{GCT1 -- summary of real images and attacks}
\label{GCT1_description}
\end{table}

Evaluations are performed following the protocol used in prior works for each dataset. Further, results for each dataset are reported using the same metrics that previous works used to reported their performance, for fair comparison. Half Total Error-Rate (HTER)~\cite{iso_jtc_sc} is reported for Idiap Replay-Attack and 3DMAD, Equal Error Rate (EER)~\cite{wen2015face} for MSU MSFD, and Attack Presentation Classification Error Rate (APCER)~\cite{iso_jtc_sc}, Bona Fide Presentation Classification Error Rate (BPCER))~\cite{iso_jtc_sc} and $\text{ACER} = (\text{APCER} + \text{BPCER}) / 2$~\cite{peng2017face} for Idiap Replay-Mobile dataset. Results of PAD performance on GCT1 are reported using APCER, BPCER, ACER, EER, and the Area Under the Receiver Operating Curve (AUC). Ablation study was performed by training and evaluating the base CNN model of RoPAD (BM) without the UAI components and results of these experiments are also reported.

\subsection{Evaluation Results}

\paragraph{Idiap Replay-Attack:} Table~\ref{Results_on_Replay-Attack} summarizes the experimental results on the Idiap Replay-Attack dataset. As shown, the proposed model achieves a perfect HTER of $0$ on this dataset, outperforming the state of the art~\cite{phan2016face}.

\begin{table}
\setlength{\tabcolsep}{1.5em} 
\centering
\begin{tabular}{lc}
\toprule
\textbf{Method} & \textbf{HTER} \\
\cmidrule(r){1-1}  \cmidrule(l){2-2}
LBP+CCoLBP~\cite{peng2018ccolbp} & 5.38 \\
CCoLBP~\cite{peng2018ccolbp} & 5.25 \\
LDP+TOP~\cite{phan2016face} & 1.75 \\
\cmidrule(r){1-1} 
BM  & 0.38 \\
RoPAD  & \textbf{0}\\
\bottomrule
\end{tabular}
\caption{Test HTER (\%) on Idiap Replay-Attack} 
\label{Results_on_Replay-Attack}
\end{table}

\begin{table}
\centering
\setlength{\tabcolsep}{0.75em} 
\begin{tabular}{lccc}
\toprule
\textbf{Method} & \textbf{ACER} & \textbf{APCER} & \textbf{BPCER}\\
\cmidrule(r){1-1}  \cmidrule(l){2-4}
IQM~\cite{costa2016replay} & 13.64  & 19.87  & 7.40\\
Gabor~\cite{costa2016replay} & 9.53  & 7.91  & 11.15 \\
LBP+GS-LBP~\cite{peng2017face} & 1.74  & 2.09 &1.38 \\
LGBP~\cite{peng2017face} & 1.50 & 2.08 &0.91 \\
LGBP (video)~\cite{peng2017face} & 1.25  & 1.40 &1.10 \\
\cmidrule(r){1-1} 
BM  & 0.90 & \textbf{0} & 1.80 \\
RoPAD  & \textbf{0} & \textbf{0} & \textbf{0} \\
\bottomrule
\end{tabular}
\caption{Test results (\%) on Idiap Replay-Mobile.} 
\label{Results_on_Replay-Mobile}
\end{table}

\begin{figure}[b]
\centering
\includegraphics[width=\linewidth,trim={0.5cm 0.35cm 0.5cm 0.5cm},clip]{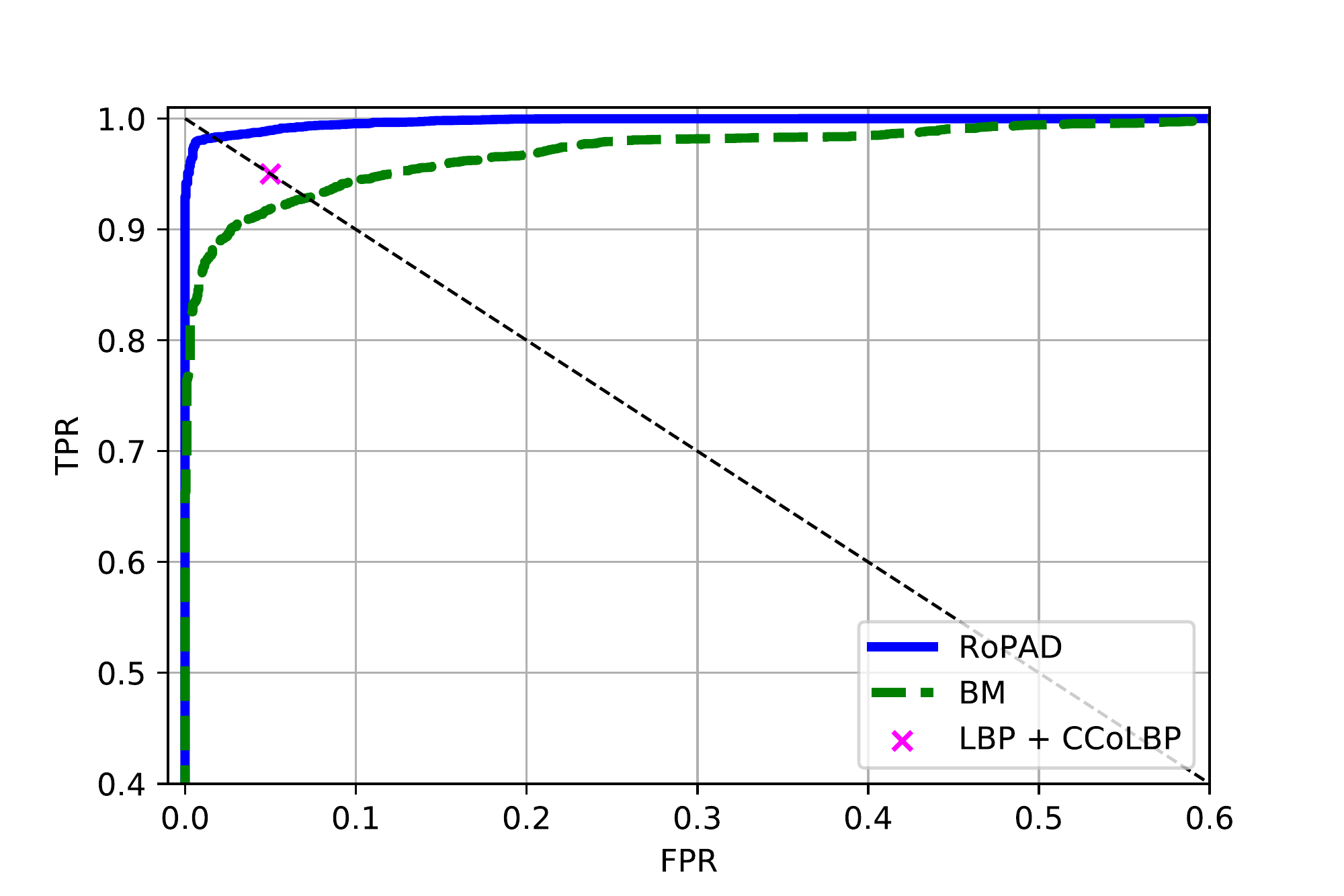}
\caption{Receiver Operating Curves for MSU MSFD}
\label{MFSD_RoC}
\end{figure}

\paragraph{Idiap Replay-Mobile:} In Table~\ref{Results_on_Replay-Mobile} we summarize the results of our experiments on the Idiap Replay-Mobile dataset. While BM outperforms the previous state-of-the-art on the ACER and APCER scores, the proposed RoPAD model performs the best on all metrics, achieving a perfect score of $0$ on each.

\paragraph{MSU MFSD:} The proposed RoPAD outperforms the previous state-of-the-art models on the MSU MSFD dataset, as shown in Table~\ref{Results_on_MFSD}. In contrast, BM performs significantly worse than the previous best model. This large performance boost is, hence, credited to the UAI component of RoPAD, and is further highlighted by the Receiver Operating Curve (ROC) shown in Figure~\ref{MFSD_RoC}.

\paragraph{3DMAD:} Table~\ref{3DMAD_random} summarizes results of our experiments on the 3DMAD dataset. While the ablation version BM performs worse than previous state-of-the-art, the proposed RoPAD achieves a perfect HTER of $0$ on this dataset also.

\begin{table}
\setlength{\tabcolsep}{1.5em} 
\centering
\begin{tabular}{lc}
\toprule
\textbf{Method} & \textbf{EER} \\
\cmidrule(r){1-1}  \cmidrule(l){2-2}
DoG-LBP+SVM \cite{wen2015face} & 23.10 \\
LBP + SVM\cite{wen2015face} & 14.70 \\
IDA + SVM \cite{wen2015face}& 8.85 \\
LDP + TOP \cite{phan2016face} & 6.54 \\
CCoLBP\cite{peng2018ccolbp} & 5.83 \\
LBP + CCoLBP \cite{peng2018ccolbp} & 5.00 \\
\cmidrule(r){1-1} 
BM & 7.15\\
RoPAD & \textbf{1.70}\\
\bottomrule
\end{tabular}
\caption{Test EER (\%) on MFSD} 
\label{Results_on_MFSD}
\end{table}

\begin{table}
\centering
\setlength{\tabcolsep}{1.5em} 
\begin{tabular}{lc}
\toprule
\textbf{Method} & \textbf{HTER} \\
\cmidrule(r){1-1}  \cmidrule(l){2-2}
LBP + LDA \cite{Erdogmus2014Spoofing} & 0.95 \\
\cmidrule(r){1-1}
BM  & 1.00\\
RoPAD & \textbf{0}\\
\bottomrule
\end{tabular}
\caption{Test HTER (\%) on 3DMAD} 
\label{3DMAD_random}
\end{table}

\begin{figure}[b]
\centering
\includegraphics[width=\linewidth,trim={0.5cm 0.35cm 0.5cm 0.5cm},clip]{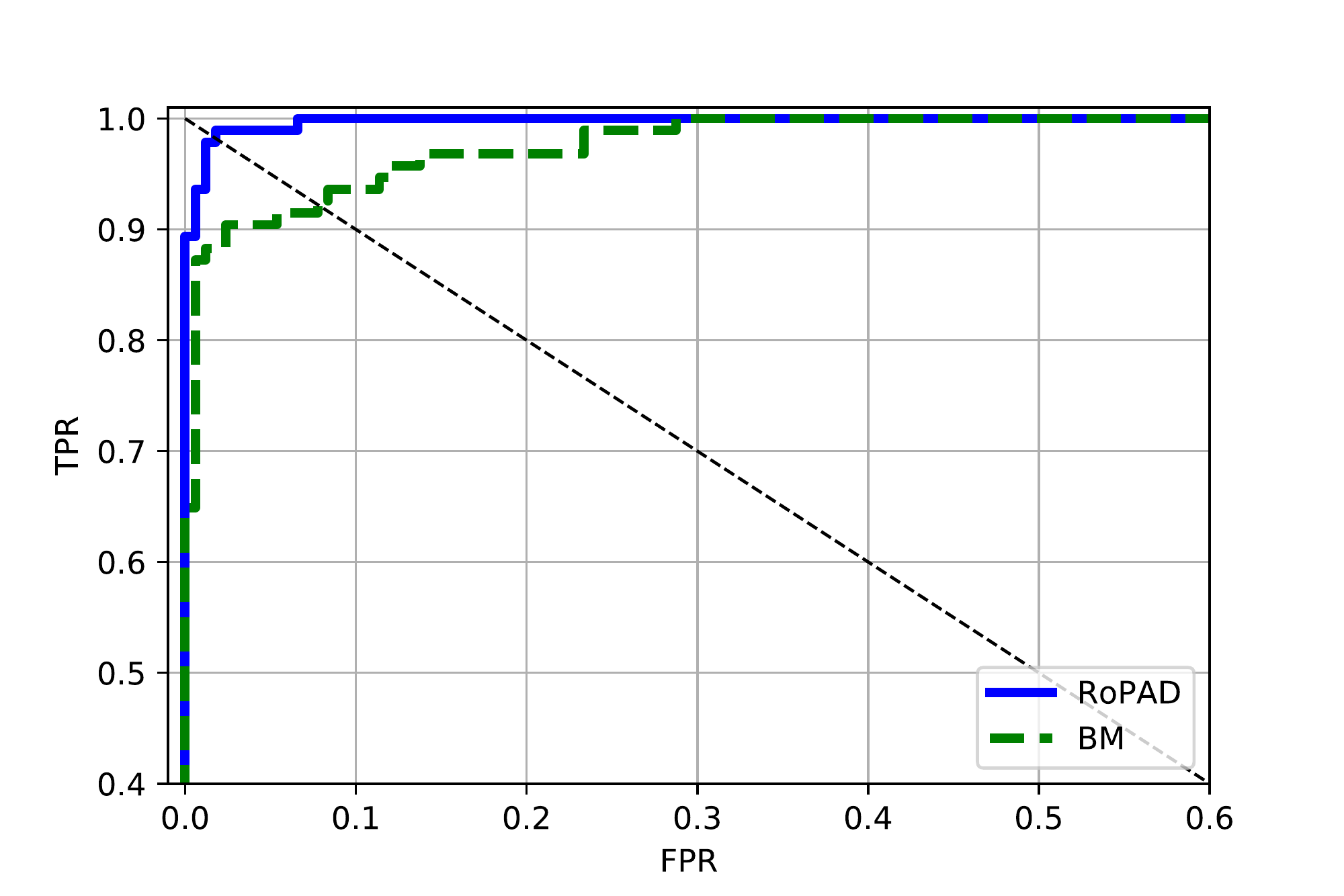}
\caption{Receiver Operating Curves for GCT1}
\label{GCT1_RoC}
\end{figure}

\begin{table*}
\setlength{\tabcolsep}{1.5em} 
\centering
\begin{tabular}{lcccccc}
\toprule
\textbf{Method} &\textbf{APCER} &\textbf{BPCER} &\textbf{ACER} &\textbf{EER} &\textbf{Validation AUC} &\textbf{Test AUC} \\
\cmidrule(r){1-1}  \cmidrule(l){2-7}   
BM       & 28.7 & \textbf{0.5} & 14.6 & 7.4 & 0.957 & 0.983 \\ 
RoPAD       & \textbf{4.2} & 1.0 & \textbf{2.5} & \textbf{1.8} & \textbf{1.000} & \textbf{0.998} \\
\bottomrule
\end{tabular}
\caption{Results on GCT1 --- all metrics except AUC are reported as percentages (\%)}
\label{GCT1_result}
\end{table*}


\paragraph{GCT1:} The proposed model achieves near perfect score at PAD on the GCT1 dataset, as shown in Table~\ref{GCT1_result}. In comparison to the base model, the ACER, EER and AUC of RoPAD are 12.1, 5.6 and 1.5 percentage points higher, respectively, which highlights a clear improvement in performance due to the incorporation of UAI in RoPAD, as further reflected in Figure~\ref{GCT1_RoC}.

In summary, taking into consideration the results of the proposed model on the aforementioned benchmark datasets, it is evident that the proposed RoPAD outperforms previous state-of-the-art models across the board. Results of the ablation version of the proposed model, on the other hand, show that the UAI component of RoPAD is crucial to achieving this outstanding performance.

%% file: sections/05_conclusions.tex
\section{Conclusion and Future Work}
\label{sec:conclusion}

The increasing demand for face-based biometric authentication calls for robust technologies that are protected against attacks that can fool such systems. In this paper, we presented a novel deep neural network model, RoPAD, for detecting presentation attacks in such systems. RoPAD is designed as a deep convolutional neural network and trained to make robust predictions by employing unsupervised adversarial invariance, which makes RoPAD invariant to factors in face images that are irrelevant for presentation attack detection. Results of extensive experimental evaluation on several datasets show that RoPAD achieves state-of-the-art performance at presentation attack detection.

The base CNN model of the proposed RoPAD is inspired from the VGG16 model. In future work, we plan to experiment with more sophisticated models like ResNet~\cite{He_2016_CVPR} to create improved versions of RoPAD.

%% file: main.bbl
\begin{thebibliography}{10}\itemsep=-1pt

\bibitem{atoum2017face}
Y.~Atoum, Y.~Liu, A.~Jourabloo, and X.~Liu.
\newblock Face anti-spoofing using patch and depth-based cnns.
\newblock In {\em Biometrics (IJCB), 2017 IEEE International Joint Conference
  on}, pages 319--328. IEEE, 2017.

\bibitem{boulkenafet2015face}
Z.~Boulkenafet, J.~Komulainen, and A.~Hadid.
\newblock Face anti-spoofing based on color texture analysis.
\newblock In {\em Image Processing (ICIP), 2015 IEEE International Conference
  on}, pages 2636--2640. IEEE, 2015.

\bibitem{boulkenafet2017face}
Z.~Boulkenafet, J.~Komulainen, and A.~Hadid.
\newblock Face antispoofing using speeded-up robust features and fisher vector
  encoding.
\newblock {\em IEEE Signal Processing Letters}, 24(2):141--145, 2017.

\bibitem{chingovska2012effectiveness}
I.~Chingovska, A.~Anjos, and S.~Marcel.
\newblock On the effectiveness of local binary patterns in face anti-spoofing.
\newblock In {\em Proceedings of the 11th International Conference of the
  Biometrics Special Interes Group}, number EPFL-CONF-192369, 2012.

\bibitem{elu}
D.~Clevert, T.~Unterthiner, and S.~Hochreiter.
\newblock Fast and accurate deep network learning by exponential linear units
  (elus).
\newblock {\em CoRR}, abs/1511.07289, 2015.

\bibitem{costa2016replay}
A.~Costa-Pazo, S.~Bhattacharjee, E.~Vazquez-Fernandez, and S.~Marcel.
\newblock The replay-mobile face presentation-attack database.
\newblock In {\em Biometrics Special Interest Group (BIOSIG), 2016
  International Conference of the}, pages 1--7. IEEE, 2016.

\bibitem{Erdogmus2014Spoofing}
N.~Erdogmus and S.~Marcel.
\newblock Spoofing in 2d face recognition with 3d masks and anti-spoofing with
  kinect.
\newblock In {\em IEEE Sixth International Conference on Biometrics: Theory,
  Applications and Systems}, pages 1--6, 2014.

\bibitem{gan20173d}
J.~Gan, S.~Li, Y.~Zhai, and C.~Liu.
\newblock 3d convolutional neural network based on face anti-spoofing.
\newblock In {\em Multimedia and Image Processing (ICMIP), 2017 2nd
  International Conference on}, pages 1--5. IEEE, 2017.

\bibitem{Gragnaniello2015AnIO}
D.~Gragnaniello, G.~Poggi, C.~Sansone, and L.~Verdoliva.
\newblock An investigation of local descriptors for biometric spoofing
  detection.
\newblock {\em IEEE Transactions on Information Forensics and Security},
  10:849--863, 2015.

\bibitem{He_2016_CVPR}
K.~He, X.~Zhang, S.~Ren, and J.~Sun.
\newblock Deep residual learning for image recognition.
\newblock In {\em The IEEE Conference on Computer Vision and Pattern
  Recognition (CVPR)}, June 2016.

\bibitem{iso_jtc_sc}
{ISO/IEC JTC 1/SC 37 - Biometrics. Information Technology Biometric
  presentation attack detection part 1: Framework.}
\newblock Standard, International Organization for Standardization, 2016.
\newblock \url{https://www.iso.org/obp/ui/iso}.

\bibitem{jaiswal2018unsupervised}
A.~Jaiswal, Y.~Wu, W.~Abd-Almageed, and P.~Natarajan.
\newblock Unsupervised adversarial invariance.
\newblock In {\em Advances in Neural Information Processing Systems 31}, pages
  5097--5107. Curran Associates, Inc., 2018.

\bibitem{jourabloo2018face}
A.~Jourabloo, Y.~Liu, and X.~Liu.
\newblock Face de-spoofing: Anti-spoofing via noise modeling.
\newblock {\em arXiv preprint arXiv:1807.09968}, 1(2):3, 2018.

\bibitem{liu2018learning}
Y.~Liu, A.~Jourabloo, and X.~Liu.
\newblock Learning deep models for face anti-spoofing: Binary or auxiliary
  supervision.
\newblock In {\em Proceedings of the IEEE Conference on Computer Vision and
  Pattern Recognition}, pages 389--398, 2018.

\bibitem{maatta2011face}
J.~M{\"a}{\"a}tt{\"a}, A.~Hadid, and M.~Pietik{\"a}inen.
\newblock Face spoofing detection from single images using micro-texture
  analysis.
\newblock In {\em Biometrics (IJCB), 2011 international joint conference on},
  pages 1--7. IEEE, 2011.

\bibitem{nogueira2016fingerprint}
R.~F. Nogueira, R.~de~Alencar~Lotufo, and R.~C. Machado.
\newblock Fingerprint liveness detection using convolutional neural networks.
\newblock {\em IEEE Trans. Information Forensics and Security},
  11(6):1206--1213, 2016.

\bibitem{peng2017face}
F.~Peng, L.~Qin, and M.~Long.
\newblock Face presentation attack detection using guided scale texture.
\newblock {\em Multimedia Tools and Applications}, pages 1--27, 2017.

\bibitem{peng2018ccolbp}
F.~Peng, L.~Qin, and M.~Long.
\newblock Ccolbp: Chromatic co-occurrence of local binary pattern for face
  presentation attack detection.
\newblock In {\em 2018 27th International Conference on Computer Communication
  and Networks (ICCCN)}, pages 1--9. IEEE, 2018.

\bibitem{phan2016face}
Q.-T. Phan, D.-T. Dang-Nguyen, G.~Boato, and F.~G. De~Natale.
\newblock Face spoofing detection using ldp-top.
\newblock In {\em Image Processing (ICIP), 2016 IEEE International Conference
  on}, pages 404--408. IEEE, 2016.

\bibitem{raghavendra2015presentation}
R.~Raghavendra, K.~B. Raja, and C.~Busch.
\newblock Presentation attack detection for face recognition using light field
  camera.
\newblock {\em IEEE Transactions on Image Processing}, 24(3):1060--1075, 2015.

\bibitem{ramachandra2017presentation}
R.~Ramachandra and C.~Busch.
\newblock Presentation attack detection methods for face recognition systems: a
  comprehensive survey.
\newblock {\em ACM Computing Surveys (CSUR)}, 50(1):8, 2017.

\bibitem{vgg16}
K.~Simonyan and A.~Zisserman.
\newblock Very deep convolutional networks for large-scale image recognition.
\newblock {\em CoRR}, abs/1409.1556, 2014.

\bibitem{wen2015face}
D.~Wen, H.~Han, and A.~K. Jain.
\newblock Face spoof detection with image distortion analysis.
\newblock {\em IEEE Transactions on Information Forensics and Security},
  10(4):746--761, 2015.

\bibitem{yang2014learn}
J.~Yang, Z.~Lei, and S.~Z. Li.
\newblock Learn convolutional neural network for face anti-spoofing.
\newblock {\em arXiv preprint arXiv:1408.5601}, 2014.

\end{thebibliography}
